\documentclass{article}
\usepackage{ijcai21}

\usepackage{times}
\usepackage{soul}
\usepackage{url}
\usepackage[hidelinks]{hyperref}
\usepackage[utf8]{inputenc}
\usepackage[small]{caption}
\usepackage{graphicx}
\usepackage{amsmath}
\usepackage{amsthm}
\usepackage{booktabs}
\usepackage{algorithm}
\usepackage{algorithmic}
\usepackage{amsfonts,amssymb}
\usepackage{multirow}
\usepackage{subfigure}
\title{AMA-GCN: Adaptive Multi-layer Aggregation Graph Convolutional Network for Disease Prediction
}

\author{
Hao Chen$^{1,2}$\and
Fuzhen Zhuang$^{3,6}$\footnote{corresponding authors}\and
Li Xiao$^{1,2,5*}$\and
Ling Ma$^1$\and
Haiyan Liu$^4$\and

Ruifang Zhang$^4$\and
Huiqin Jiang$^{1*}$\And
Qing He$^{1,2}$\\
\affiliations
$^1$Zhengzhou University, Zhengzhou, China\\
$^2$Key Lab of Intelligent Information Processing of Chinese Academy of Sciences (CAS), Institute of Computing Technology, CAS, Beijing 100190, China\\
$^3$Institute of Artificial Intelligence, Beihang University, Beijing 100191, China\\
$^4$The First Affiliated Hospital of Zhengzhou University, Zhengzhou, China\\
$^5$Ningbo Huamei Hospital, University of the Chinese Academy of Sciences, Ningbo, China\\
$^6$Xiamen Data Intelligence Academy of ICT, CAS, China
\emails
chenhao199503@gs.zzu.edu.cn,
zhuangfuzhen@buaa.edu.cn,
andrew.lxiao@gmail.com,
ielma@zzu.edu.cn,
yanmai8023@126.com,
zhangruifang999@hotmail.com,
iehqjiang@zzu.edu.cn,
heqing@ict.ac.cn
}

\begin{document}

\maketitle

\begin{abstract}
Recently, Graph Convolutional Networks (GCNs) have proven to be a powerful mean for Computer Aided Diagnosis (CADx). This approach requires building a population graph to aggregate structural information, where the graph adjacency matrix represents the relationship between nodes. Until now, this adjacency matrix is usually defined manually based on phenotypic information. In this paper, we propose an encoder that automatically selects the appropriate phenotypic measures according to their spatial distribution, and uses the text similarity awareness mechanism to calculate the edge weights between nodes. The encoder can automatically construct the population graph using phenotypic measures which have a positive impact on the final results, and further realizes the fusion of multimodal information. In addition, a novel graph convolution network architecture using multi-layer aggregation mechanism is proposed. The structure can obtain deep structure information while suppressing over-smooth, and increase the similarity between the same type of nodes. Experimental results on two databases show that our method can significantly improve the diagnostic accuracy for Autism spectrum disorder and breast cancer, indicating its universality in leveraging multimodal data for disease prediction.
\end{abstract}

\section{Introduction}
There is a growing body of researchers that have realized the potential for graph convolutional networks in medical-related fields. Recently, GCNs have been widely used to solve a variety of medical problems already, including localization of landmarks in craniomaxillofacial \cite{2020Automatic}, bone age assessment \cite{2020Towards}, representation learning for medical fMRI images \cite{2020Spatio} and so on. In this work, we focus on Computer Aided Diagnosis (CADx) \cite{2019Graph}, which uses computer technology to assist physicians in disease prediction. At present, deep learning methods have been widely used in disease prediction tasks and achieve great results, but in order to further improve diagnostic accuracy, it is necessary to make full use of complex medical multimodal data to extract the effective information hidden in it. These data usually include medical imaging and corresponding non-imaging phenotypic measures (e.g. subject’s age, height, or acquisition site), which are usually non-Euclidean and difficult to be processed by traditional deep learning methods. Moreover, not every phenotypic measure is helpful for disease prediction and it is still a tedious and time-consuming task for people to select phenotypic measures that can have a positive effect on classification results.

Inspired by the success of GCNs in social network analysis and recommendation systems \cite{Michael2017Geometric}, graph-based methods are usually used in multimodal data processing. At present, a large number of researchers have made great contributions to apply GCNs in CADx. At first, the population graph constructed by the features extracted from the multimodal data is used as the input of GCNs \cite{2018Disease}. However, the phenotypic measures selected in this method contribute the same to the edge weights, but should actually be different. Recently, there are two methods to solve this problem, which are multi-graph fusion methods \cite{2018RNN,2019Self,2019Graph} and single comprehensive graph methods \cite{2020Edge}. Although these methods deal with the phenotypic measures differently, they both assign appropriate weights to it. For example, Huang et al. propose that an encoder can be built to calculate the connection between nodes directly using multimodal data, and the only graph constructed in the end can be used as the input of GCNs.

Although the above two methods can achieve high accuracy for disease prediction, their common limitation is that there is still no effective method to screen out phenotypic measures that have a negative impact on classification results. Moreover, the parameter quantity of multi-graph fusion methods will become very large with the increase in phenotypic measures, which seriously affects the real-time performance of the model. In addition, although existing methods, such as the JK-GCN \cite{2018Representation} or EV-GCN \cite{2020Edge}, have tried to apply layer aggregation mechanism to GCNs to learn more structural information, due to the actual medical situation being too complex, it is still challenging to determine an appropriate aggregation strategy.

In order to address the above challenges, we present a new similarity-aware adaptive calibrated multi-layer aggregation GCN structure called Adaptive Multi-layer Aggregation GCN(AMA-GCN). As shown in Figure 1, this structure contains a specially designed encoder to select effective phenotypic measures and calculate the edge weights, namely phenotypic measure selection and weight encoder (PSWE). Besides, we propose two separately designed GCN models to aggregate the deep structural information and increase the similarity between different objects of the same type, respectively. This structure can improve the accuracy of the model and its robustness. The main contributions of our work can be summarized as follows:
\begin{itemize}
\item We design \textbf{PSWE} to automatically select the best combination of phenotypic measures to interpret the similarity between subjects and calculate their scores.
\item A multi-layer aggregation graph convolutional network with multiple aggregation modes is introduced to consider more appropriate structural information for each node. And a dynamic updating mechanism is devised to increase the similarity between nodes of the same type.
\item We test AMA-GCN on real-world dataset. The results show that our method is superior to other models known at present in terms of validation set accuracy. 
\end{itemize}

\begin{figure*}[htb]
\includegraphics[width=\textwidth]{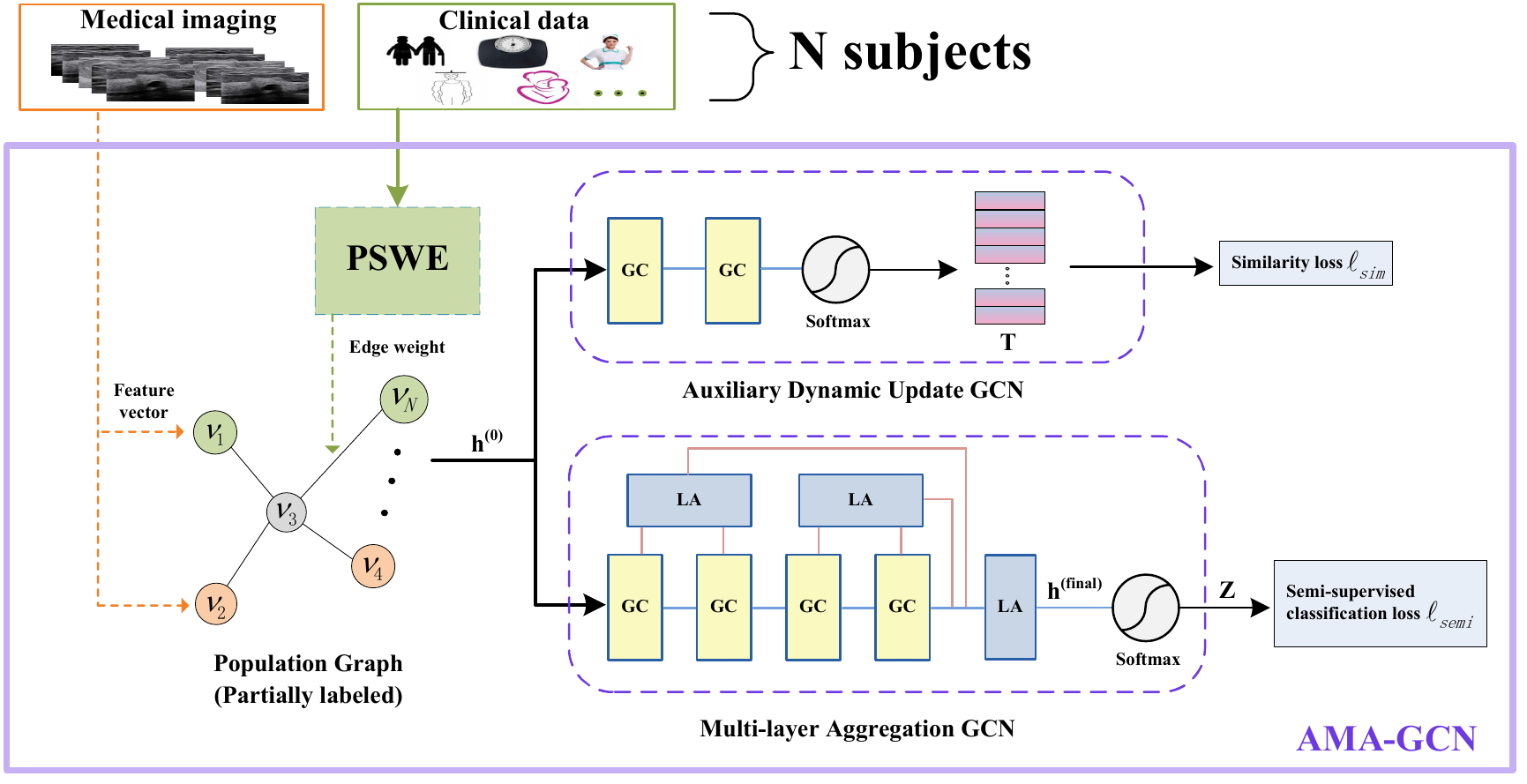}
\caption{Overall framework of the proposed method. PSWE: phenotypic measure selection and weight encoder. GC: graph convolution. LA: information aggregation layer. Colors in the graphs: green and orange - labeled diagnostic values (e.g., healthy or diseased), grey: unlabeled.} \label{fig1}
\end{figure*}

\section{Related Work}
In the past, disease classification based on deep learning was usually achieved using medical imaging. Recently, in order to further improve the classification accuracy, non-imaging phenotypic data has also been included in the research scope, that is, medical multimodal data.
\subsection{Medical Imaging Based Approach}
This method analyzes the medical imaging of patients from different perspectives, so as to obtain as many image features as possible to improve diagnostic accuracy. For example, MSE-GCN \cite{2020MSE} is proposed to extract temporal and spatial information respectively from fMRI and DTI to comprehensively analyze medical imaging. Zhang et al. have achieved great result performance for nodule-level malignancy prediction by using the transfer learning method \cite{2020trl}. This method requires that the input data must be specific types of medical imaging, while the imaging methods of different diseases are usually diverse (e.g., ultrasound imaging for the breast and fMRI for the brain), which makes this method difficult to apply to the diagnosis of other diseases. Moreover, model learning only relies on medical imaging, and does not take into account the rich non-image phenotypic data, which does not conform to the diagnostic habits of professional doctors in the actual situation.

\subsection{Multimodal Data Based Approach}
In order to further improve the classification accuracy, researchers study patients' non-imaging phenotypic data while studying medical imaging. This complex data is called multimodal data, which is usually non-Euclidean and difficult to be processed by traditional deep learning methods. At present, GCNs are usually used to process multimodal data in the medical field. This method needs to define the population graph where nodes represent patients, edges represent connections between patients, and edge weights represent similarity between patients. Specifically, phenotypic measures are usually used to calculate similarity between patients, and image features extracted from medical imaging are stored in the corresponding nodes. Furthermore, according to the different processing methods of phenotypic measures, GCNs methods can be divided into multi-graph fusion methods and single comprehensive graph methods. Multi-graph fusion methods usually construct graphs for each phenotypic measure and process them separately, and design different multi-graph fusion methods to assign weights to each phenotypic measure, such as RNN \cite{2018RNN}, self-attention \cite{2019Self}, or LSTM \cite{2019Graph}. Single comprehensive graph methods directly extract features from the multimodal data to build a comprehensive graph as input. For example, Huang et al. \cite{2020Edge} propose to use an encoder to process multimodal data. The above methods have achieved great results, but there are still some limitations: not every phenotypic measure has a positive effect on classification results. In fact, although our proposed method is also to build a single comprehensive graph to predict disease, it does not need to determine the appropriate phenotypic measures through a large number of experiments as in the past, but is automatically completed by the proposed encoder.

\section{Methodology}
\subsection{Preliminaries}
In our study, the population graph is defined as an undirected graph $ G=(V,E,A) $, where $V$ is the set of vertices and each vertex represents a patient; $E$ denotes the set of edges; $A$ denotes the adjacency matrix of population graph $G$, whose elements are the edge weights. As shown in Figure 1, the adjacency matrix $A$ is obtained by using the proposed PSWE, which can automatically find the appropriate phenotypic measures and calculate the corresponding phenotypic measure selection scores (PMS-scores), and then calculate the edge weights. Note that traditional GCNs usually artificially select phenotypic measures.

The node feature matrix $X$ is also the input of our model. We define $X$ as $X=(x_1,x_2,...,x_n) \in \mathbb{R}^{n \times m}$, where $n$ denotes the number of samples, $m$ represents the dimension of the features. We select feature extracted from medical imaging of the two datasets involved in our experiment as the node information. For the ABIDE database, we use functional connectivity derived from resting-state functional Magnetic Resonance Imaging (rs-fMRI) for Autism Spectrum Disorder (ASD) classification. Rudie et al. \cite{2013Altered} proposed that ASD is linked to disruptions in the functional and structural organization of the brain, Abraham et al. \cite{2016Deriving} further demonstrated this assertion. More accurately, we use the vectorized functional connectivity matrices as feature vectors. For our collection of breast cancer mass ultrasound image data, we use ResNet-50 to extract features of medical imaging and directly verify the performance of these features on ResNet-50 and Ridge classifier.

\subsection{Phenotypic Measure Selection and Weight Encoder}
Edges are the channels through which the node filters information from its neighbors and represent relationships between nodes. Our hypothesis is that some appropriate non-imaging phenotypic data can provide critical information as complement to explain the associations between subjects. Therefore, selecting the best combination of phenotypic measures to interpret the similarity between subjects and assigning appropriate weights to them are the keys to our experiment. In our studies, it will be done by PSWE, and the implementation process of PSWE as shown in Figure 2.

Considering a set of $H$ non-imaging phenotypic measures $K=\{K_h\}$, including quantitative (e.g. subject’s age, height, or BMI) and non-quantitative (e.g. subject’s calcification or capillary distribution) phenotypic measures. The population graph’s adjacency matrix $A$ is defined as follows:
\begin{equation}
A(v,w)=\sum_{h=1}^H \alpha_h * \gamma(K_h (v),K_h (w))
\end{equation}
where $\alpha_h$ is the PMS-scores of phenotypic measure $K_h$; $\gamma$ is a measure of the distance between the value of phenotypic measure $K_h$ of two graph nodes. $\alpha_h$ is defined as follows:
\begin{equation}
\alpha_h =
\begin{cases}
H * \frac{n^{K_h}}{\sum_{h=1}^H n^{K_h}}, & {H * n^{K_h} \ge \sum\limits_{h=1}^H n^{K_h},} \\
0, & {otherwise.}
\end{cases}
\end{equation}
where $n^{K_h}$ is a measure of the number of samples in which phenotypic measure $K_h$ meets the requirements, and it is defined differently depending on the type of phenotypic measures. When $K_h$ is a non-quantitative phenotypic measure (e.g. calcification or edema), we define $n^{K_h}$ as a function with respect to a threshold $\theta$:
\begin{equation}
n^{K_h} =
\begin{cases}
\frac{1}{U}\sum\limits_{u=1}^U \sum\limits_{p=1}^P n_{p\_u}^{K_h}, & {\frac{n_{u}^{K_h}-n_{p\_u}^{K_h}}{n_{p\_u}^{K_h}} <\theta,}\\
0, & {otherwise.}
\end{cases}
\end{equation}
where $n_{p\_u}^{K_h}$ is a measure of the number of samples with the value $u$ and category $p$ in the phenotypic measure $K_h$; $n_{u}^{K_h}$ is a measure of the number of samples with the value $u$ in the phenotypic measure $K_h$. Meanwhile, when $K_h$ is a quantitative phenotypic measure (e.g. subject age or BMI), we define $n^{K_h}$ as a function with respect to a threshold $\delta$:
\begin{equation}
n^{K_h} =
\begin{cases}
\sum\limits_{p=1}^P n_{p\_s}^{K_h}, & {\frac{n_{p}^{K_h}-n_{p\_s}^{K_h}}{n_{p\_s}^{K_h}} <\delta,}\\
0, & {otherwise.}
\end{cases}
\end{equation}
where $n_{p}^{K_h}$ is a measure of the number of samples with the category $p$ in the phenotypic measure $K_h$; We define a closed interval $\mathcal D$ from $\alpha$ to $\beta$, where $\mathcal D\in value(K_h)$, and $n_{p\_s}^{K_h}$ is a measure of the number of samples with the category $p$ and value $s\notin \mathcal D$ in the phenotypic measure $K_h$.

\begin{figure}
\centering
\includegraphics[width=8cm]{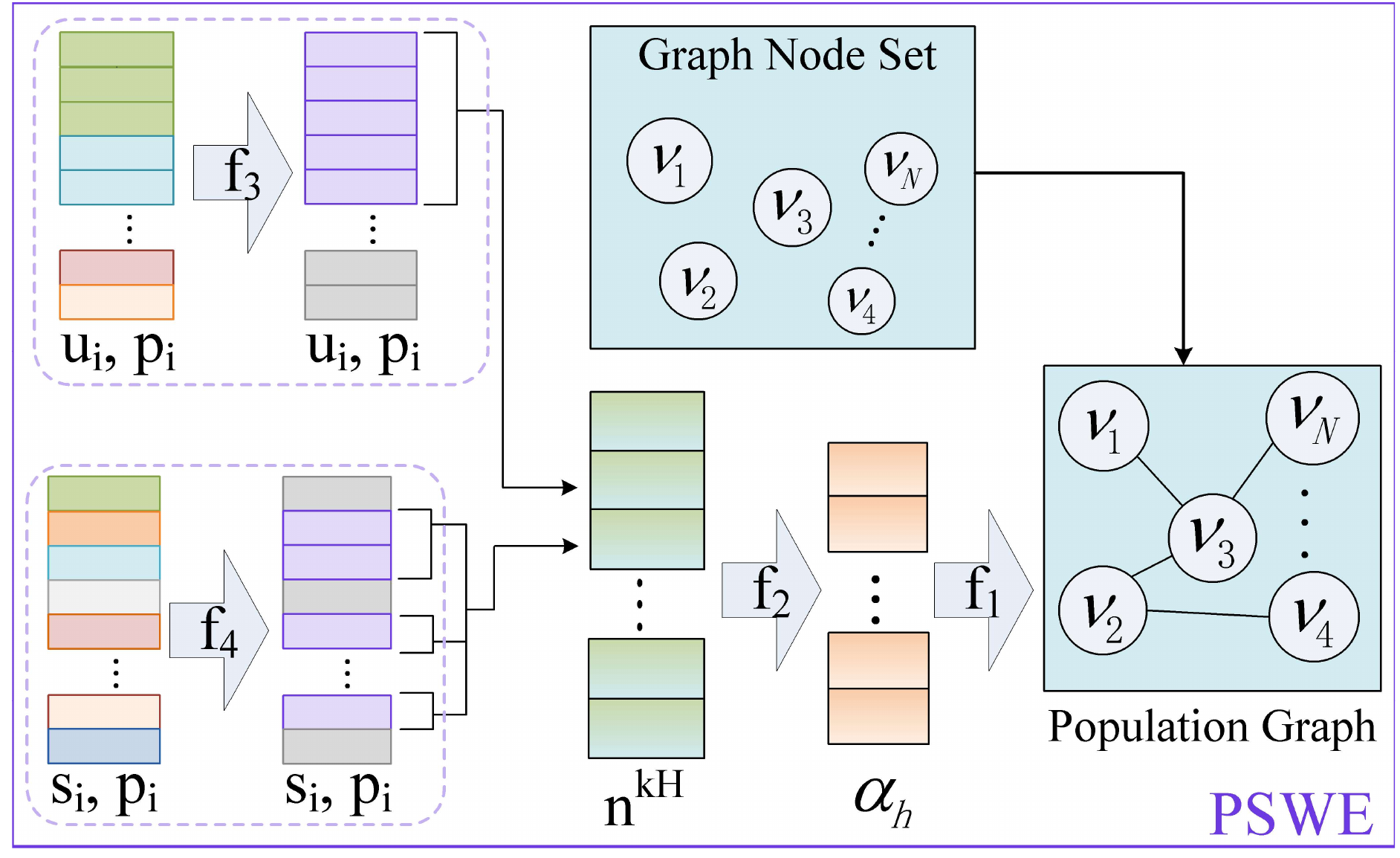}
\caption{Overview of the proposed PSWE. $u_i$: values of non-quantitative phenotypic measures corresponding to subject $i$. $s_i$: values of quantitative phenotypic measures corresponding to subject $i$. $p_i$: the category corresponding to subject $i$.} \label{fig2}
\end{figure}

$\gamma$ is also defined differently depending on the type of phenotypic measure. For non-quantitative phenotypic measures such as family history, we define $\gamma$ as the Kronecker delta function, meaning that the similarity highly between two subjects if their values of phenotypic measure $K_h$ are the same. For quantitative phenotypic measures such as subject’s age, we define $\gamma$ as a function with respect to a closed interval $\mathcal D$ from $\alpha$ to $\beta$, where $\mathcal D\in value(K_h)$:
\begin{equation}
\begin{aligned}
& \gamma(K_h (v),K_h (w)) = \\
& \begin{cases}
1, & {K_h (v),K_h (w) \notin \mathcal D,}\\
\frac{1}{e^{\sqrt[3]{\lvert K_h (v)-K_h (w) \rvert}}}, & {\lvert K_h (v)-K_h (w) \rvert < \beta-\alpha,}\\
0, & {otherwise.}
\end{cases}
\end{aligned}
\end{equation}

The influence of effective phenotypic measures selected by PSWE on the classification results will be investigated in our experiments, so as to visually demonstrate the performance of PSWE.

\subsection{Model Structure Design}
In order to make nodes from different dense blocks to obtain sufficient effective information on each layer while suppressing over-smooth, we reconstruct the architecture of the network with multi-layer aggregation mechanism so that the information from different layers can be adaptively fused into the final expression of the node. As shown in Figure 1, in order to get the key features of structural information, the first two aggregation layers that directly integrate the information from graph convolution layers are aggregated in the way of maxPooling. Besides, the final aggregation layer uses the way of concating to fully summary the information aggregated by each layer. The rebuilt model enables each node to automatically integrate the appropriate information.

The input graph data $h^{(0)}$ is processed by the above model, and the output $h^{(final)}$ is then calculated by the softmax activation function. The final output $Z\in \mathbb{R}^{n \times P}$denotes the label prediction for all data in which each row $Z_i$ denotes the label prediction for the $i-th$ node. The optimal weight matrices trained by minimizing the cross-entropy loss function as:
\begin{equation}
\mathcal L_{semi} =- \sum_{i\in L} \sum_{j=1}^P Y_{ij}\ln{Z_{ij}}
\end{equation}
where $L$ indicates the set of labeled nodes. $Y_{ij}$ represents the label information of the data.

In addition, we introduce an auxiliary dynamic update GCN to increase the similarity between nodes of the same types while encouraging competition. As shown in Figure 1, this model's output is also calculated by the softmax activation function. The final output $T\in \mathbb{R}^{n\times P}$ denotes the classification score for all data in which each row $T_i$ denotes the classification score for the $i-th$ node. Note that $T$ is the output of the auxiliary model, which is different from the output $Z$ of the semi-supervised classification model. The similarity loss function as:
\begin{equation}
\mathcal L_{sim} = \tanh{\frac{{\sum_{i\in L} \sum_{j=1}^P (Y_{ij}-T_{ij})^2 + \xi}}{2\sigma^2}}
\end{equation}
where $\xi$ is a minimal constant; $\sigma$ determines the width of the kernel.

Then, the joint representation is used to compute a fusion loss. It poses extra regularization that can help generalization \cite{2020GREEN}. The final loss function is defined as:
\begin{equation}
\mathcal L = \mathcal L_{semi} + \lambda \mathcal L_{sim}
\end{equation}
where $\mathcal L_{semi}$ and $\mathcal L_{sim}$ are defined in Eq.(6) and Eq.(7), respectively. Parameter $\lambda \ge 0$ is a tradeoff parameter with a default value of 1.

\section{Experiments and Results}

\subsection{Dataset}
To verify the effectiveness of our model, we evaluate it on the Autism Brain Imaging Data Exchange (ABIDE) database \cite{2014ABIDE}. The ABIDE publicly shares fMRI and the corresponding phenotypic data (e.g., age and gender) of 1112 subjects, and notes whether these subjects have Autism Spectrum Disorder (ASD). In order to compare fairly with state-of-the-art \cite{2020Edge} on the ABIDE, we select the same 871 subjects consisting of 403 normal and 468 ASD individuals, and perform the same data preprocessing steps \cite{2020Edge,2018Disease}. Then we delete the subjects with empty values, and finally select the 759 subjects consisting of 367 normal and 392 ASD individuals. Besides, approved by the local Institutional Review Board, the local hospital provides ultrasound images of breast nodules and the corresponding phenotypic data (e.g., age, gender, and calcification) of 572 sets, and follows the diagnostic results given by the radiologist for each subject to note whether these subjects have Breast Cancer. These data are acquired from 121 different patients consisting of 55 adenopathy and 66 breast cancer individuals, and each set contains a total of six images (two ultrasound static imaging and four ultrasound elastography). We remove cases without complete phenotypic measures and adjust all images to 256×256, and then organize the collected data into the breast cancer detection dataset (BCD) and verify the universality of proposed model on it.

\begin{table}[]
\centering
\begin{tabular}{lc}
\hline
Hyperparameter   description & Value  \\ \hline
Layer number of the MLA-GCN  & 5      \\
Layer number of the ADU-GCN  & 2      \\
Chebyshev polynomial         & 3      \\
Number of node features      & 2000   \\
Graph convolution kernel     & 16     \\
Learning rate of MLA-GCN     & 0.005  \\
Learning rate of ADU-GCN     & 0.05   \\
Regularization parameter     & 0.0005 \\
Dropout probability          & 0.3    \\
Number of training epoch     & 300    \\
Tradeoff parameter $\lambda$ & 1      \\
Optimizer                    & Adam   \\ \hline
\end{tabular}
\caption{Experiment hyperparameter setting. MLA-GCN: multi-layer aggregation GCN. ADU-GCN: auxiliary dynamic update GCN.}
\label{tab1}
\end{table}

\subsection{Baseline Methods and Settings}
We compare the AMA-GCN with the following baselines:\\
\textbf{ResNet-50} \cite{2016Identity}: A single modality classification approach using only images.\\
\textbf{Ridge classifier} \cite{2016Deriving}: A single modality classification approach using only features extracted from medical imaging data.\\
\textbf{GCN} \cite{2018Disease}: A model extracting features contained in medical multimodal data, which is usually used to deal with non-Euclidean data.\\
\textbf{JK-GCN} \cite{2018Representation}: A model using an aggregation layer before the output layer to integrate information.\\
\textbf{GLCN} \cite{2020Semi}: A model using graph learning mechanism to constantly optimize graph structure to improve the classification effect.\\
\textbf{EV-GCN} \cite{2020Edge}: A model using a custom encoder to obtain the association between nodes from non-imaging phenotypic data, and JK-GCN is used to aggregate information.\\
\textbf{EV-GCN+PS}: A model using the phenotypic measures selected by proposed PSWE as the basis for constructing the population graph, and using EV-GCN to extract structural information.
 
\subsubsection{Ablation Study}
To investigate how the PSWE, the multi-layer aggregation mechanism and similarity loss function improve the performance of the proposed model, we conduct the ablation study on the following variants of AMA-GCN:\\
\textbf{AMA-GCN(noP)} is a model that uses the same phenotypic measures with baseline methods to build the population graph as input and uses proposed model for training.\\
\textbf{AMA-GCN(noW)} is a model that only uses PSWE to select effective phenotypic measures, and uses a two-layer GCN for training. Note that the weight ratio calculated by PSWE is not used.\\
\textbf{AMA-GCN(noA)} is a model that uses PSWE to build the population graph, but uses a two-layer GCN for training.\\
\textbf{AMA-GCN(noS)} is a model without the similarity loss function. 

In order to ensure a fair comparison, when we do not use PSWE to select the appropriate phenotypic measures, we choose gender and acquisition sites as the basis for constructing the population graph of ABIDE database, and choose age as the basis for constructing the population graph of BCD dataset. This setting applies to all baseline models. The hyperparameters of the experiment are shown in Table 1. We employ 10-fold cross-validation to evaluate the performance of the model and implement our model using TensorFlow. In order to evaluate the performance of models, we choose overall accuracy (ACC) and area under the curve (AUC) as the evaluation indicators.

\subsection{Results and Analysis}
\begin{table}[]
\begin{center}
\begin{tabular}{l|llll}
\hline
\multirow{2}{*}{} & \multicolumn{2}{c}{ABIDE} & \multicolumn{2}{c}{BCD} \\ \cline{2-5} 
                 & ACC         & AUC        & ACC           & AUC        \\ \hline
ResNet-50        & 0.626       & 0.679      & 0.897         & 0.955          \\
Ridge classifier & 0.636       & 0.688      & 0.901         & 0.961          \\
GCN              & 0.705       & 0.731      & 0.916         & 0.950          \\
JK-GCN           & 0.722       & 0.736      & 0.947         & 0.972          \\
GLCN             & 0.707       & 0.725      & 0.932         & 0.965          \\
EV-GCN           & 0.829       & 0.876      & 0.967         & 0.987          \\
EV-GCN+PS        & 0.936       & 0.949      & 0.971         & 0.989          \\
\textbf{AMA-GCN}          & \textbf{0.984}       & \textbf{0.983}      & \textbf{0.994}         & \textbf{0.998}          \\ \hline
AMA-GCN(noP)     & 0.724       & 0.747      & 0.956         & 0.979          \\
AMA-GCN(noW)     & 0.955       & 0.974      & 0.974         & 0.987          \\
AMA-GCN(noA)     & 0.972       & 0.986      & 0.976         & 0.983          \\
AMA-GCN(noS)     & 0.981       & 0.981      & 0.990         & 0.992          \\ \hline
\end{tabular}
\end{center}
\caption{ Quantitative comparisons between different methods on ABIDE and BCD.}
\label{tab2}
\end{table}

We compare our AMA-GCN with the six baseline methods for predictive classification of disease on the ABIDE database and BCD dataset, as shown in Table 2. It can be observed that single-mode models (i.e. ResNet-50 and Ridge classifier) only use medical imaging data as the basis for classification, and their overall performance is poor. Comparatively, graph-based methods (GCN, JK-GCN, GLCN, EV-GCN and ours) yield larger performance gains, benefiting from analyzing associations between nodes in the population graphs. The proposed method, AMA-GCN, obtains an average accuracy of 98.4$\%$ and 99.4$\%$ on ABIDE and BCD datasets, respectively, outperforming the recent SoTA method EV-GCN, which employs an adaptive population graph with variational edges and uses JK-GCN to aggregate structure information. We notice that the performance of graph-based methods is highly sensitive to the phenotypic measures used to construct graphs, where the phenotypic measures $K_h=\{$gender, acquisition sites$\}$ on ABIDE database used by EV-GCN yields an average accuracy of 82.9$\%$. To prove the effectiveness of the phenotypic measures selected by our PSWE, we train EV-GCN using the same phenotypic measures as those in our model. As depicted in Table 2, it results in 10.7$\%$ accuracy and 7.3$\%$ AUC improvement on ABIDE database respectively, indicating that the appropriate phenotypic measures are indeed the key to improving the performance of disease prediction. Meanwhile, the universality in leveraging multimodal data for disease prediction of the proposed AMA-GCN architecture is relatively validated according to the comparison results on BCD dataset in Table 2.

\begin{figure}
\centering
\subfigure[ACC of the ABIDE dataset]{
        \includegraphics[width=0.22\textwidth,height=3cm]{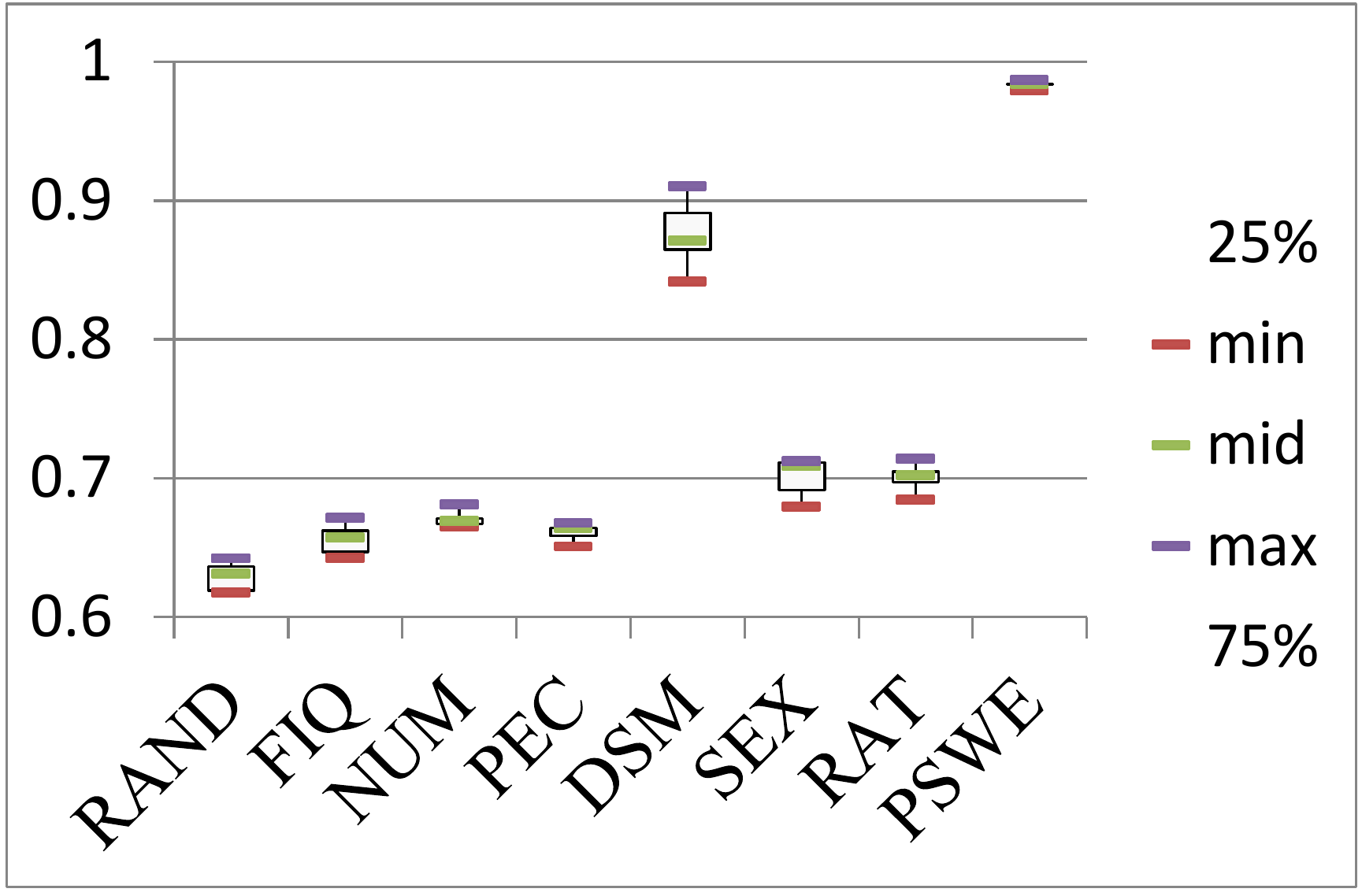}
    }
\subfigure[AUC of the ABIDE dataset]{
        \includegraphics[width=0.22\textwidth,height=3cm]{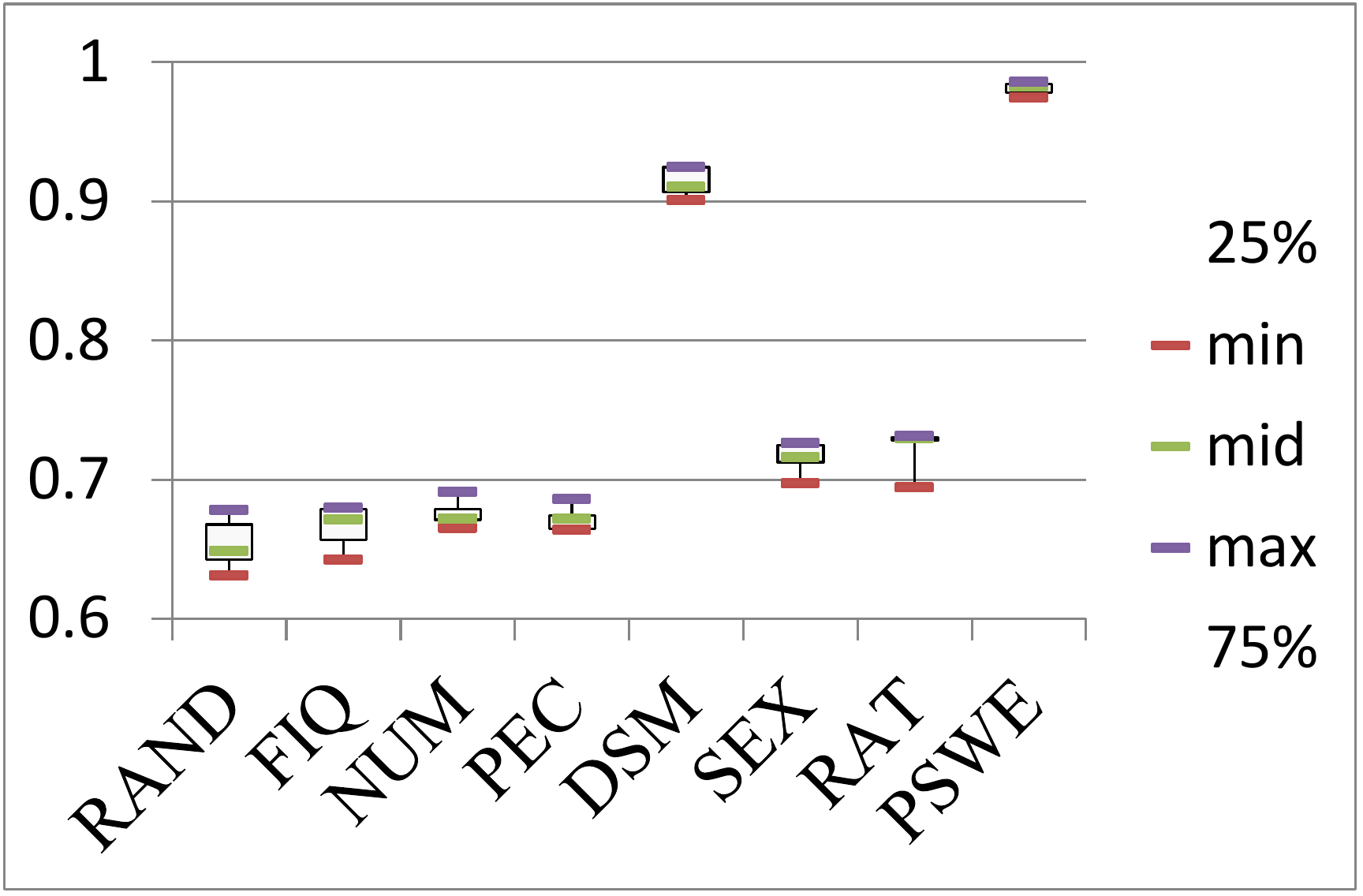}
    }
\subfigure[ACC of the BCD dataset]{
        \includegraphics[width=0.22\textwidth,height=3cm]{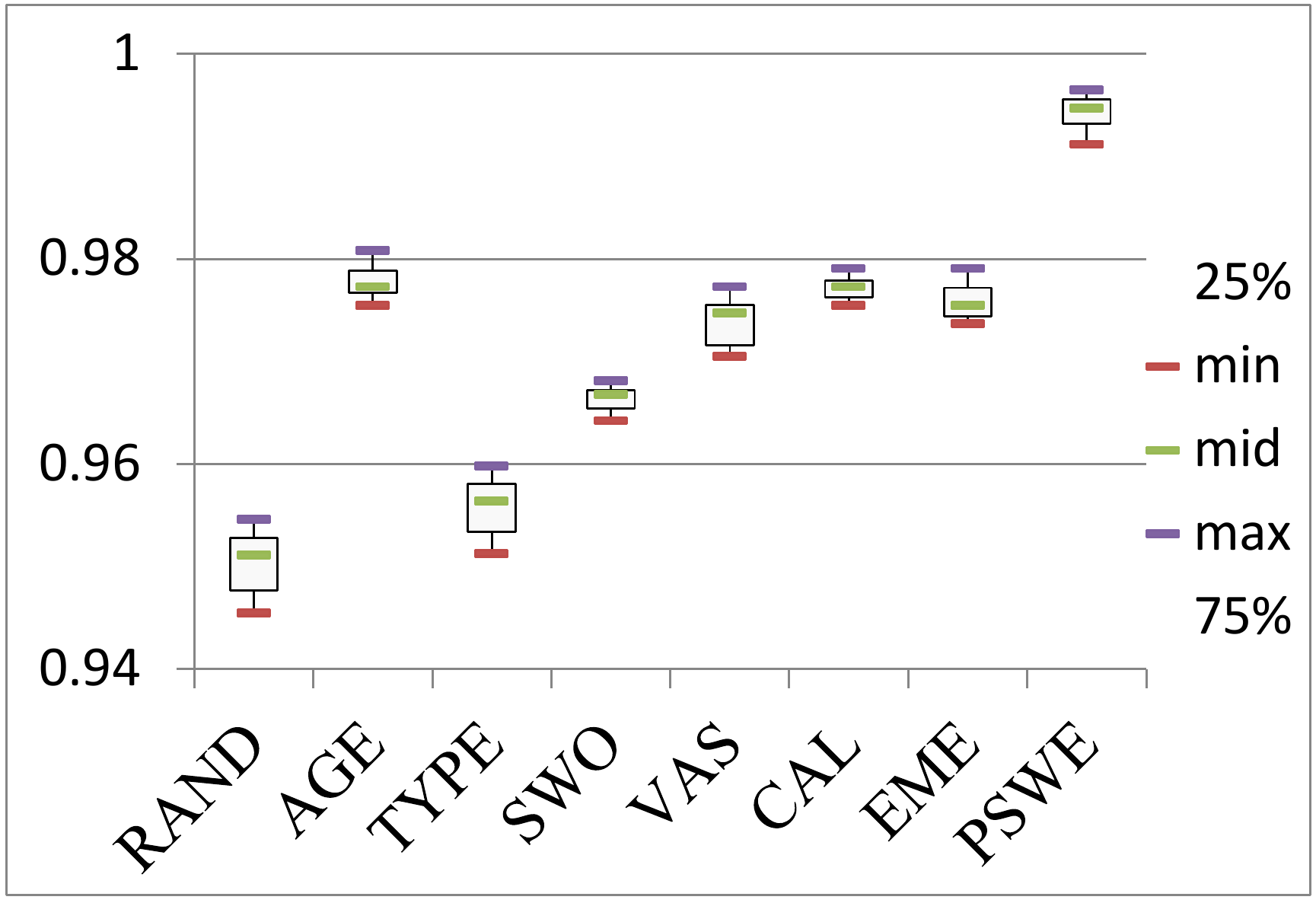}
    }
\subfigure[AUC of the BCD dataset]{
        \includegraphics[width=0.22\textwidth,height=3cm]{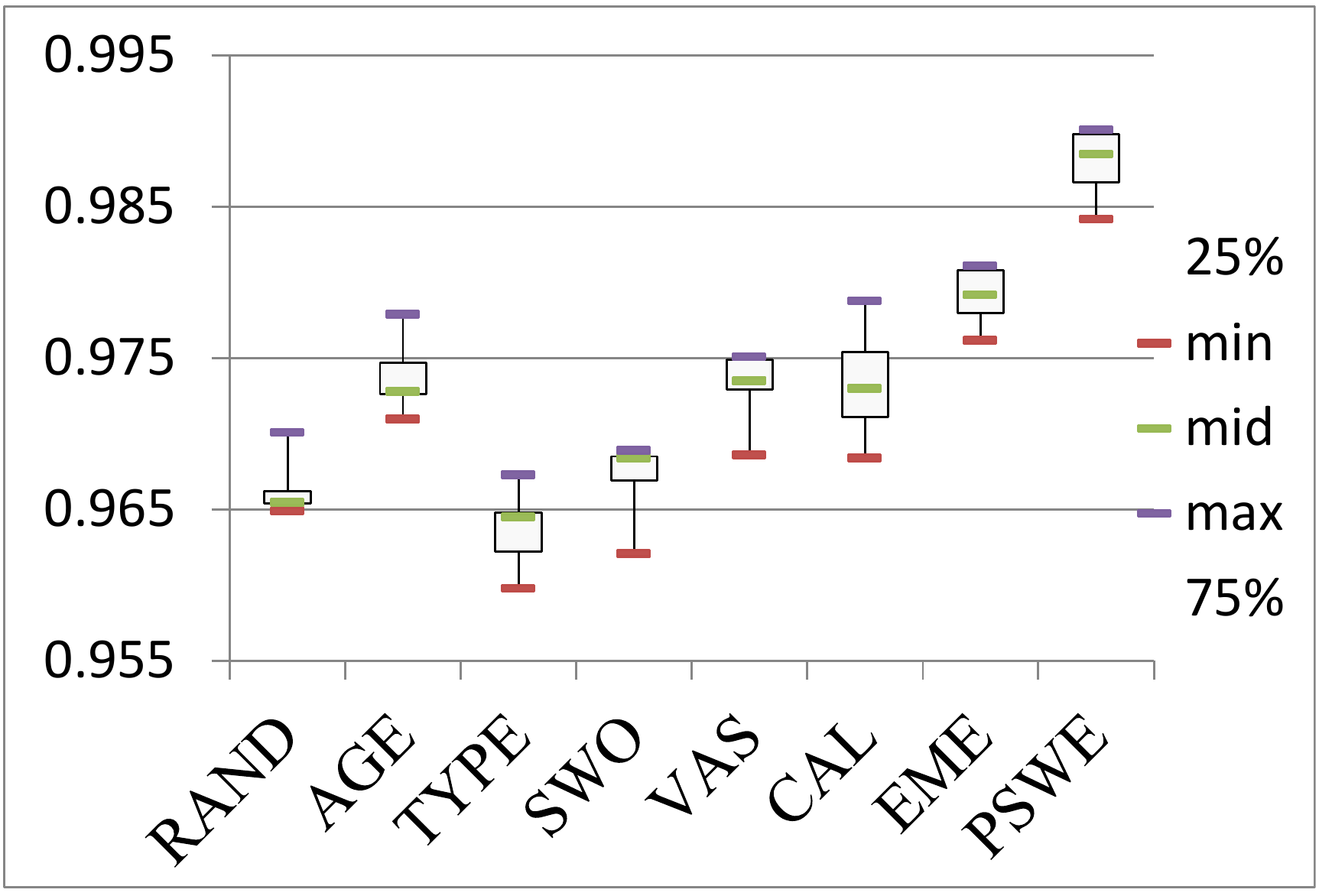}
    }
\caption{Effects of effective phenotypic measures selected by PSWE on results, the evaluation indicators are Accuracy (ACC) and Area Under Curve(AUC).} 
\label{fig3}
\end{figure}

\subsubsection{Ablation Experiments}
As an ablation study, we test whether removing the PSWE, the multi-layer aggregation mechanism or similarity loss function affect model performance, and the results as shown in Table 2. It can be observed that the phenotypic measures selected by PSWE can significantly improve the classification accuracy. Using only these phenotypic measures without considering PMS-scores, the average accuracy of the constructed graphs on the ABIDE and BCD datasets is improved by 25.0$\%$ and 5.8$\%$, respectively. The classification accuracy is further improved by 1.7$\%$ and 0.2$\%$ on ABIDE and BCD datasets respectively after calculating PMS-scores to the selected phenotypic measures. Although the latter can also improve the classification accuracy, the improvement is far less than the former, which indicates that finding the appropriate phenotypic measures is the key to disease prediction. In addition, the effectiveness of multi-layer aggregation mechanism in improving classification accuracy is also validated according to the comparison results in Table 2.

\subsection{Phenotypic Measures Analysis}
In order to demonstrate our conclusion that finding the appropriate phenotypic measures is the key to solving the problem of disease classification, and also to prove the effectiveness of proposed PSWE, we further explore the impact of each phenotypic measure on the classification results. As shown in Figure 3, using the single phenotypic measure selected by PSWE to construct the population graph as input, their performances on ABIDE and BCD datasets are respectively more than 3$\%$ and 1$\%$ better than that of randomly constructed graph. Comparatively, as shown in Figure 4, using phenotypic measures not selected by PSWE as the basis for constructing graphs, the final performance is basically the same as that of randomly generated graphs, and even worse than the latter, which indicates that these phenotypic measures selected by PSWE are reasonable. Meanwhile, PSWE achieves the best performance by combining multiple effective phenotypic measures, and its accuracy on ABIDE database is even 33.8$\%$ higher than that of randomly constructed graph. This indicates that there is hidden complementary information among phenotypic measures, and how to learn these complementary information is the key to disease prediction.

\begin{figure}
\centering
\subfigure[ACC of the ABIDE dataset]{
        \includegraphics[width=0.22\textwidth,height=3cm]{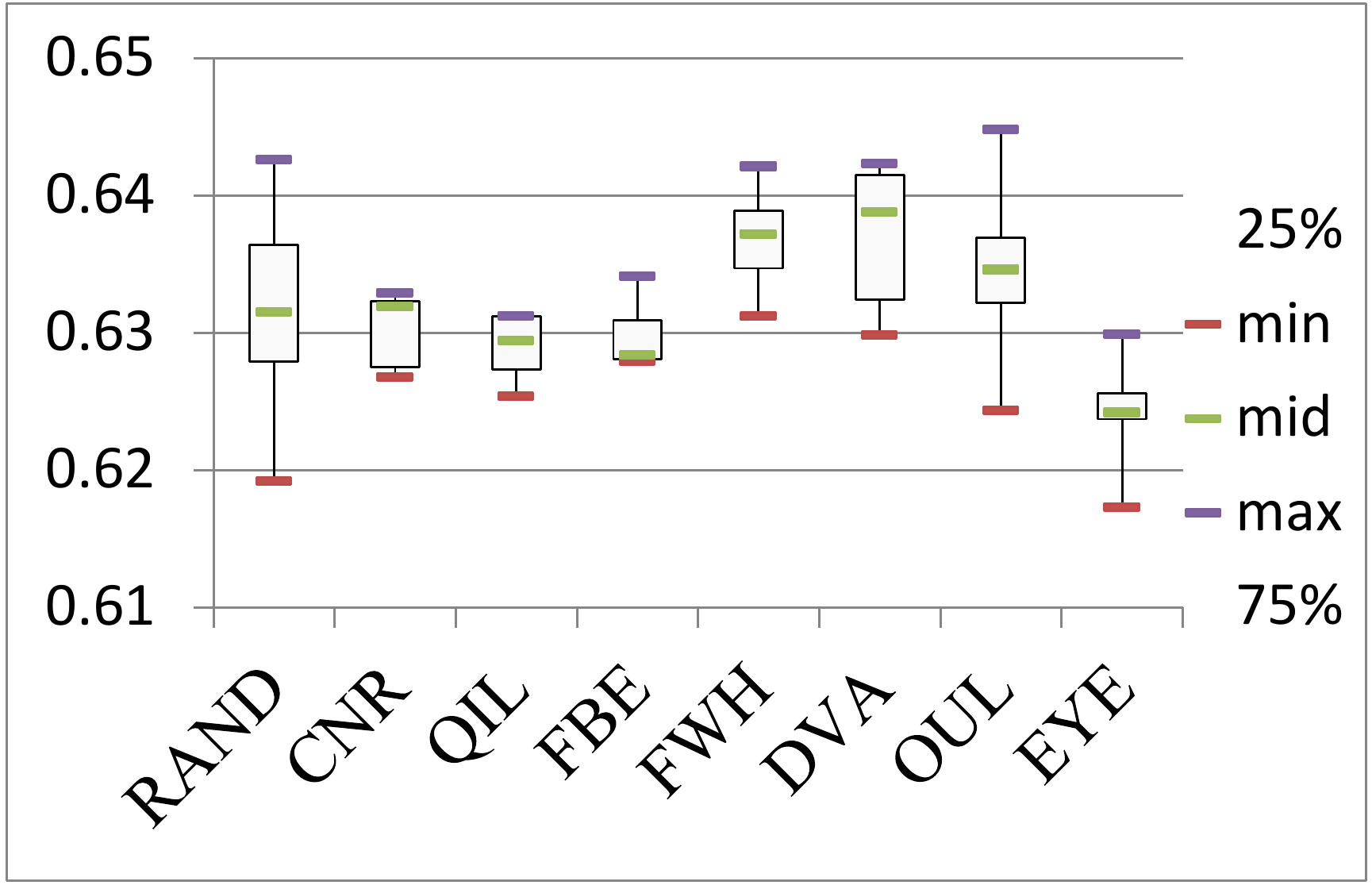}
    }
\subfigure[ACC of the BCD dataset]{
        \includegraphics[width=0.22\textwidth,height=3cm]{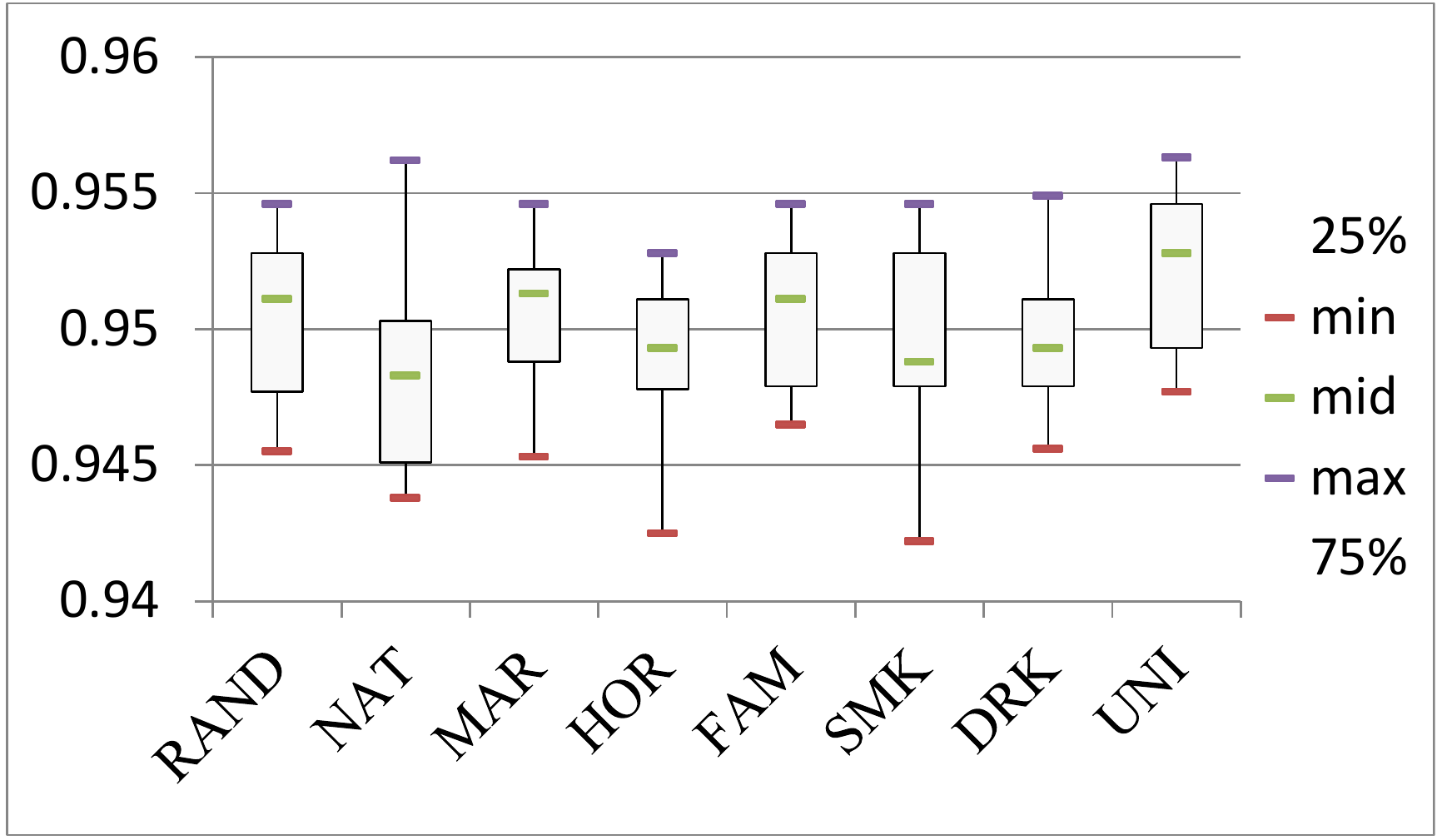}
    }
\caption{Effects of phenotypic measures not selected by PSWE on results, the evaluation indicators is Accuracy (ACC).}
\label{fig4}
\end{figure}

\section{Conclusion}
In this paper, we have proposed a generalizable graph-convolutional framework that combines multimodal data to predict disease. We designed the population graph structure according to the spatial distribution and text similarity of phenotypic measures, while allowing each effective phenotypic measure to contribute to the final prediction. We reconstructed the graph convolution model by using the multi-layer aggregation mechanism to automatically find the optimal feature information from each layer while suppressing over-smooth, and introduce another channel to increase the similarity between different objects in the same type. Experimental results show that the proposed method achieves superior performance on brain analysis and breast cancer prediction. We believe that such an extensible method can have a better use of helping people with medical multimodal data for clinical computer-aided diagnosis.

\section*{Acknowledgments}
This work was supported by the National Natural Science Foundation of China (31900979,U1604262), CCF-Tencent Open Fund, the National Key Research and Development Program of China under Grant (2017YFB1002104), the National Natural Science Foundation of China under Grant (U1811461), and Zhengzhou Collaborative Innovation Major Project (20XTZX11020).


\bibliographystyle{named}

\end{document}